\documentclass[10pt,twocolumn,letterpaper]{article}
\usepackage{cvpr}              

\usepackage{graphicx}
\usepackage{amsmath}
\usepackage{amssymb}
\usepackage{booktabs}
\usepackage{blindtext}
\usepackage{caption}
\usepackage{algorithm}
\usepackage{algorithmic}
\usepackage[switch]{lineno}
\usepackage{amsmath}
\usepackage{amssymb}
\usepackage{multirow}
\usepackage{booktabs}
\usepackage{pifont}%
\newcommand{\cmark}{\ding{51}}%
\newcommand{\xmark}{\ding{55}}%
\usepackage[pagebackref,breaklinks,colorlinks]{hyperref}

\usepackage[capitalize]{cleveref}
\crefname{section}{Section}{Secs.}
\Crefname{section}{Section}{Sections}
\Crefname{table}{Table}{Tables}
\crefname{table}{Table}{Tabs.}
\def\X{$\mathcal{X}$-Trans2Cap }

\def\cvprPaperID{1993} 
\def\confName{CVPR}
\def\confYear{2022}

\title{$\mathcal{X}$-Trans2Cap: Cross-Modal Knowledge Transfer using Transformer\\for 3D Dense Captioning}

\author{Zhihao Yuan$^{1, \dagger}$, Xu Yan$^{1, \dagger}$, Yinghong Liao$^{1}$,  Yao Guo$^{2}$, Guanbin Li$^{3}$, Shuguang Cui$^{1}$,  Zhen Li$^{1,}$\thanks{{ Corresponding author: Zhen Li. $^\dagger$ Equal first authorship.}} \\
	\\
	$^{1}$The Chinese University of Hong Kong (Shenzhen),  
	The Future Network of Intelligence Institute, \\
	Shenzhen Research Institute of Big Data, 
	$^{2}$Shanghai Jiao Tong University, 
	$^{3}$Sun Yat-sen University \\
	{\tt\small	\{{zhihaoyuan@link.}, xuyan1@link., {lizhen@}\}cuhk.edu.cn}}

\begin{document}
	
	\maketitle
	
	\begin{abstract}
		3D dense captioning aims to describe individual objects in 3D scenes by natural language, where 3D scenes are usually represented as RGB-D scans or point clouds.
		However, only exploiting single modal information, \eg, point cloud, previous approaches fail to produce faithful descriptions. 
		Though aggregating 2D features into point clouds may be beneficial, it introduces an extra computational burden, especially in the inference phase.
		In this study, we investigate a \textbf{cross}-modal knowledge {\textbf{trans}}fer using {\textbf{Trans}}former for 3D dense {\textbf{cap}}tioning, namely \textbf{X-Trans2Cap}.
		Our proposed X-Trans2Cap effectively boost the performance of single-modal 3D captioning through the knowledge distillation enabled by a teacher-student framework.
		In practice, during the training phase, the teacher network exploits auxiliary 2D modality and guides the student network that only takes point clouds as input through the feature consistency constraints.
		Owing to the well-designed cross-modal feature fusion module and the feature alignment in the training phase, X-Trans2Cap acquires rich appearance information embedded in 2D images with ease.
		Thus, a more faithful caption can be generated only using point clouds during the inference.
		Qualitative and quantitative results confirm that X-Trans2Cap outperforms previous state-of-the-art by a large margin, \ie, about {\textbf{+21}} and {\textbf{+16}} CIDEr points on ScanRefer and Nr3D datasets, respectively.
		
	\end{abstract}
	
	\section{Introduction}
	
	Hitherto, the computer vision community has witnessed significant progress in image captioning~\cite{vinyals2015show,xu2015show,karpathy2015deep,lu2017knowing,anderson2018bottom} and dense captioning~\cite{johnson2016densecap,karpathy2015deep,kim2019dense,li2019learning} under the success of deep learning techniques.
	%
	%
	Unlike image captioning describing a 2D image with a single sentence, dense captioning (DC) better interprets ``A picture is worth a thousand words".
	That is to say, for DC task, each object in an image is first perceived, then is provided more customized and detailed descriptions according to its nature and context.
	%

	\begin{figure}[t]
		
		\centering\includegraphics[width=0.95\columnwidth]{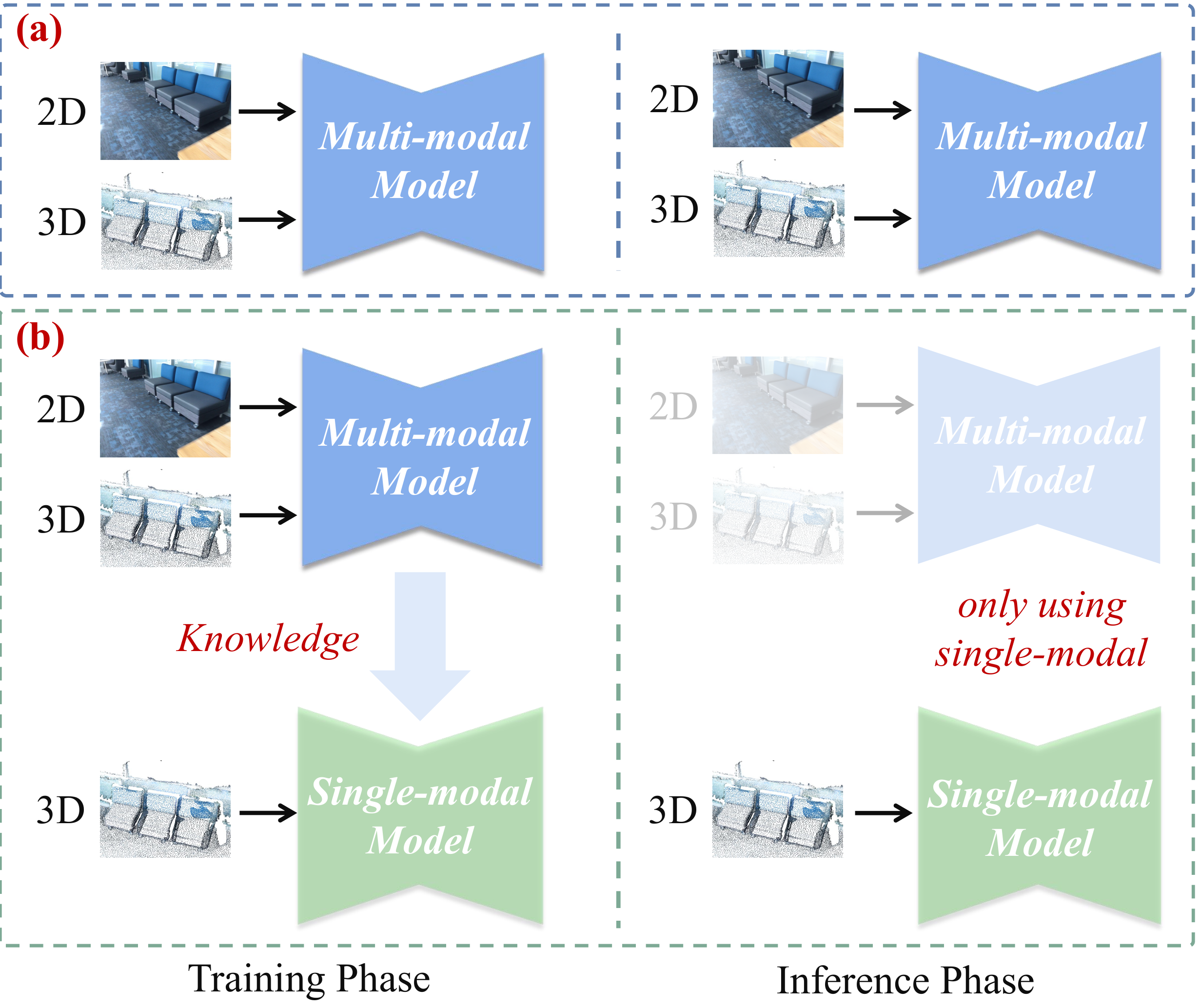}
		
		\caption{\textbf{The motivation of cross-modal knowledge transfer.} 
			(a) Previous methods use the extra 2D modality as the input in both training and inference phases. 
			(b) In contrast, we exploit a teacher-student framework with multi-modal data during training. For inference, the student network only takes 3D modality input.}
		\label{fig:fig1}
		
	\end{figure}
	
	Most recently, 3D cross-modal learning in vision and language has gained an increasing amount of interest as well.
	Several datasets~\cite{chen2020scanrefer,achlioptas2020referit3d, goyal2020rel3d} and downstream applications~\cite{huang2021text,yuan2021instancerefer} are proposed and investigated.
	Unlike 2D images with regular grids and dense pixels, 3D data represented by a set of points are unordered and scattered in the 3D space, impeding the direct extension of 2D-based methods to 3D scenarios.
	To perform dense captioning on 3D point clouds, \cite{chen2021scan2cap} proposes the first method, namely Scan2Cap, by directly combining 3D object detection with natural language generation.
	Specifically, Scan2Cap first employs a detection backbone to obtain object proposals, and then applies a relational graph and a context-aware attention captioning module to learn object relations and generate tokens.
	Besides, multi-view features extracted by the pre-trained E-Net~\cite{paszke2016enet} are further projected onto the input point cloud to enhance final captioning.
	However, Scan2Cap still has several issues:
	\textbf{1)} The object representations in Scan2Cap are defective since they are solely learned from sparse 3D point clouds, thus failing to provide strong texture and color information compared with the ones generated from 2D images.
	\textbf{2)} It requires the extra 2D input in both training and inference phases, as shown in Figure~\ref{fig:fig1} (a). However, the extra 2D information is usually computation-intensive and unavailable during inference. 
	For instance, a model training with both 2D and 3D inputs cannot apply to LiDAR scenarios that only contains 3D point cloud.
	%
	
	To address the above issues, we explore how to ease the barrier of cross-modal learning on 2D and 3D data, and investigate how to effectively combine the merits of both modalities for 3D dense captioning in this paper.
	To this end, we first time present a flexible and novel cross-modal framework, namely \textbf{X-Trans2Cap}\footnote{\url{https://github.com/CurryYuan/X-Trans2Cap}}, which transfers color and texture-aware information from 2D image into 3D object representation using Transformer~\cite{vaswani2017attention}.
	Concretely, all the instances in a given scene can be firstly extracted by 3D object detection.
	Subsequently, the 3D features of each instance and its 2D counterpart are processed by a teacher-student framework. 
	Within this framework, the teacher network takes the multi-modal inputs, while the student one only leverages the 3D inputs.
	Considering different modalities for teacher and student streams, we innovatively design a Transformer-based knowledge transfer framework with more flexible input control and better representation.
	Moreover, to further enhance the knowledge transfer, a modified knowledge distillation operation with cross-modal fusion (CMF) module and cross-modal feature alignment objective is proposed for knowledge generalization.
	Owing to the end-to-end training scheme, the priors in the 2D modality can inherently improve the teacher network and the student as well, \ie, our model takes advantage of the color and texture aware 2D representation and reduces the extra computational cost. 
	Therefore, in the inference phase, {X-Trans2Cap} can perform superior captioning performance with only 3D inputs, as shown in Figure~\ref{fig:fig1}~(b). 
	
	Sufficient experiments evaluated on the ScanRefer~\cite{chen2020scanrefer} and Nr3D~\cite{achlioptas2020referit3d} datasets have demonstrated the effectiveness of our proposed {X-Trans2Cap}.
	In specific, with the extra 2D priors and the novel framework design, {X-Trans2Cap} can effectively learn a better 3D object representation and boost the performance over the model without 2D priors, \ie, improving the CIDEr points on ScanRefer from 75.75 to 87.09.
	This result also exceeds the previous state-of-the-art Scan2Cap by about \textbf{21 CIDEr}.
	%
	
	In summary, our main contributions are threefold: 
	\begin{itemize}
		\setlength{\itemsep}{0pt}
		\setlength{\parsep}{0pt}
		\setlength{\parskip}{0pt}
		\item We first time propose \textbf{X-Trans2Cap}, a simple but effective cross-modal knowledge transfer framework for 3D dense captioning, in which an enhanced 3D representation with 2D priors is achieved.
		\item X-Trans2Cap leverages a modified knowledge distillation method through a novel cross-modal fusion module and feature alignment techniques merged in Transformer, eliminating extra computation burdens during inference while achieving superior knowledge transfer. 
		\item Our {X-Trans2Cap} gains significant performance boost on the ScanRefer~\cite{chen2020scanrefer} ({\textbf{+21.0}} CIDEr) and Nr3D~\cite{achlioptas2020referit3d} ({\textbf{+16.7}} CIDEr) datasets.
	\end{itemize}
	
	
	\begin{figure*}[t]
		\begin{center}
			
			\includegraphics[width=0.9\linewidth]{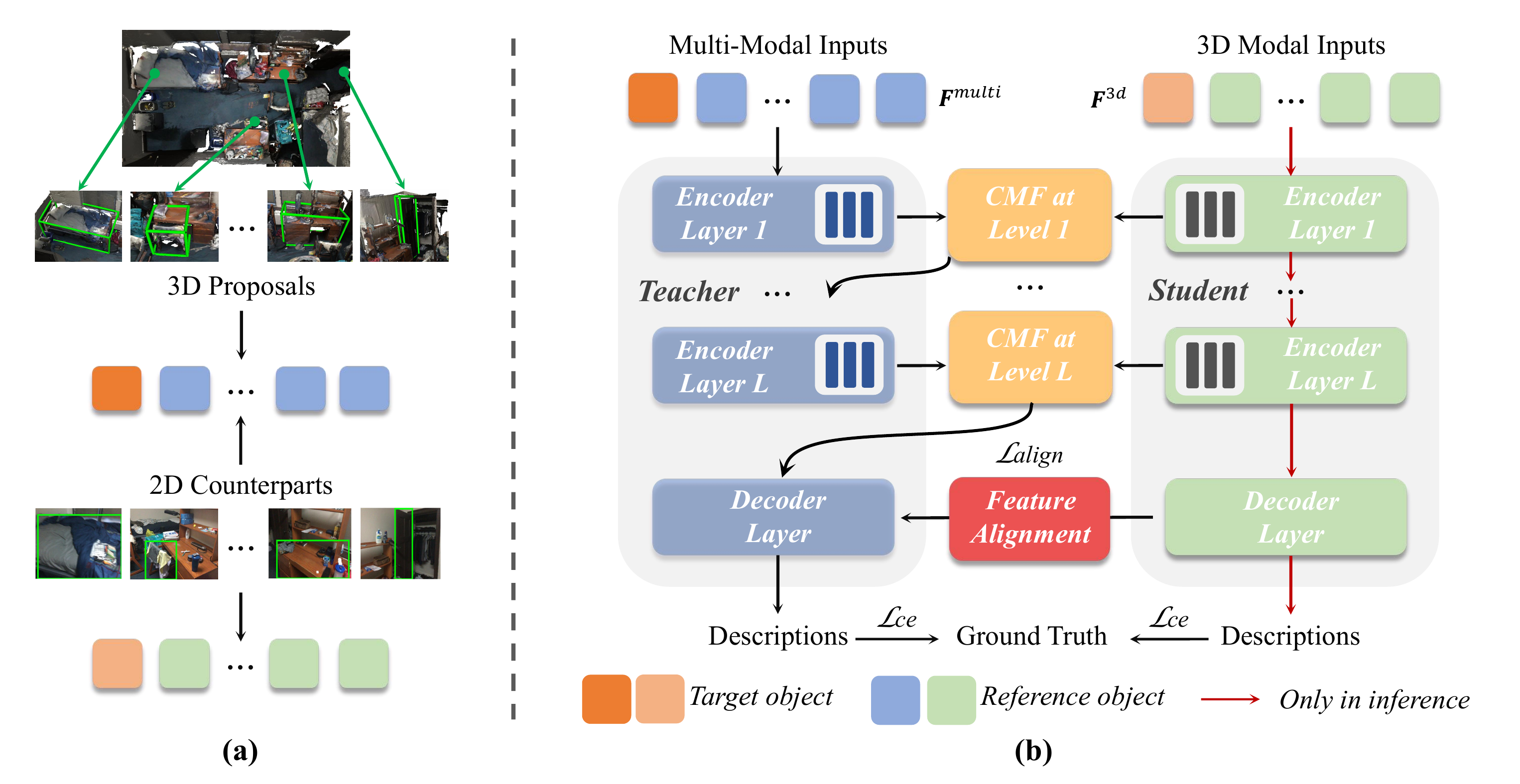}
			\vspace{-.3cm}
			\caption{\textbf{X-Trans2Cap.} 
				Part (a) describes the object representation, where the 3D proposals are utilized in the generation of pure 3D modal inputs. By integrating their 2D counterparts, multi-modal inputs are obtained.
				Part (b) illustrates the architecture of X-Trans2Cap, where the framework adheres to a teacher-student design. Both the teacher and student networks consist of several Transformer encoder layers and a decoder layer.
				%
			}
			\label{fig:fig2}
		\end{center}	
	\end{figure*}

	\section{Related Work}
	\subsection{Image Captioning and Dense Captioning} 
	A broad collection of methods have been proposed in the
	field of image captioning in the last few years~\cite{xu2015show, vinyals2015show,donahue2015long,li2017deep,lu2017knowing}. 
	Recently, many methods focus on utilizing the attention mechanism to capture meaningful information in the image, \eg, over grid regions~\cite{xu2015show,lu2017knowing} and detected objects~\cite{anderson2018bottom,lu2018neural}.
	Furthermore, some works attempt to combine attention with graph neural networks~\cite{gao2018image,yang2019auto,yao2018exploring,acl2022hu} or Transformer~\cite{cornia2020meshed} to boost performance.
	
	For the dense captioning task, it needs to generate captions for all the detected objects.
	Johnson~\etal~\cite{johnson2016densecap} is the pioneer in this challenging field.
	Along this line, \cite{yang2017dense} considers the context outside the salient image regions and takes advantage of global image features.
	\cite{kim2019dense} further introduces the object relations among detected regions.
	However, due to the limited views of a single image, the performance of image-based dense captioning methods is significantly degraded when directly transferred to 3D scenarios.

	\subsection{3D Vision and Language}
	
	Compared to image and language comprehension, 3D vision and language understanding is a relatively emerging research field. 
	Existed works focus on using language to confine individual objects, \eg, detecting referred 3D objects~\cite{chen2018text2shape} or distinguishing objects according to language phrases~\cite{achlioptas2019shapeglot}.
	Recently, ScanRefer~\cite{chen2020scanrefer} and ReferIt3D~\cite{achlioptas2020referit3d} introduce a task of localizing objects within a 3D scene given the linguistic descriptions, namely 3D visual grounding.
	TGNN~\cite{huang2021text} and InstanceRefer~\cite{yuan2021instancerefer} follow the above settings and exploit panoptic segmentation to reduce the number of proposals.
	3D dense captioning is proposed very lately in Scan2Cap~\cite{chen2021scan2cap}.
	It focuses on decomposing 3D scenes and describing the chromatic and spatial information of the objects.
	Very recently, \cite{zhenyu2021d3net} combines the above task of 3D grounding and caption to mutually enhance the performance of two tasks.
	%
	Though promising, it only takes point clouds as input to generate instance features. Compared with the well-organized 2D images containing stronger texture and color information, such representation inherently challenges the learning process.

	\subsection{Cross-modal Knowledge Transferring}
	
	Previous studies apply 2D images as the extra inputs to 3D tasks, \eg, 3D object detection~\cite{qi2018frustum,xu2018pointfusion,lahoud20172d,qi2020imvotenet}, semantic segmentation~\cite{jaritz2019multi,dai20183dmv,hu2021bidirectional} and object tracking~\cite{zheng2021box,zheng2022beyond}. 
	However, they require extra 2D information in both the training and inference phases.
	Thus, it inevitably augments computational burdens during evaluation and severely limits the efficiency in real-world applications.
	The concept of knowledge distillation was first shown by Hinton \textit{et al.}~\cite{hinton2015distilling}. Subsequent research~\cite{ba2013deep,chen2017learning} enhanced distillation by matching intermediate representations in the networks along with	outputs using different approaches.  Zagoruyko \textit{et al.}~\cite{zagoruyko2016paying} proposed to align attentional activation maps between networks. Srinivas and Fleuret~\cite{srinivas2018knowledge} improved it by applying Jacobian matching to networks. 
	In recent years, cross-modal knowledge distillation~\cite{gupta2016cross,wang2019efficient,yuan2018rgb,zhao2020knowledge} extended knowledge distillation by applying it to transferring knowledge across different modalities. 
	Very recently, there are works attempting to only utilize 2D images during training phase to address the above problems. 
	Among them, the 2D-assisted pre-training~\cite{liu2021learning}, inflating 2D convolution kernels to 3D~\cite{xu2021image2point} and joint training with mask attention~\cite{yang2021sat} are proposed.
	Unlike those, we adopt a well-designed teacher-student framework with cross-modal fusion for more efficacious knowledge transfer, and the experiment results also demonstrate that our approach is much better than previous knowledge transferring.

	\section{Method}

	Our $\mathcal{X}$-Trans2Cap is developed upon a teacher-student framework~\cite{hinton2015distilling}, which is widely exploited in the knowledge distillation research field.
	The detailed architecture of $\mathcal{X}$-Trans2Caps is presented in Figure~\ref{fig:fig2}.
	$\mathcal{X}$-Trans2Cap takes two types of features as input, \ie, pure 3D modal input for student and multi-modal input for teacher respectively.
	We first introduce the details of the above feature representation in Section.~\ref{sec:feat}.
	Then we propose a baseline model for 3D dense captioning through Transformers~\cite{vaswani2017attention} in Section.~\ref{sec:transformer}, named TransCap.
	In Section~\ref{sec:interaction}, we illustrate how $\mathcal{X}$-Trans2Cap transfers the 2D priors to the 3D representations, in which a cross-modal fusion (CMF) module is proposed.
	The details of training objectives are presented in Section~\ref{sec:loss}.
	Finally, by incorporating the above components in a whole architecture, we illustrate the data flow of $\mathcal{X}$-Trans2Cap in training and inference phases in Section~\ref{sec:training}.
	
	\subsection{Object Representation}
	\label{sec:feat}
	
	As shown in Figure~\ref{fig:fig2} (a), our framework takes object-level representation as input, and each object feature is denoted as a token.
	Given that there are $M$ objects in the 3D scene, in the remaining section, the objects set is represented as $\textbf{O}=\{O_m\}_{m=1}^M$, in which $O_m$ and $O_m^{att}$ are depicted as the $m$-th object and the attribute of the $m$-th object, respectively.
	In each iteration, we randomly choose an object as the target object ($O_*$) to be described as in \cite{chen2021scan2cap}. 
	The other $M-1$ objects, \ie, $\{O_m \in \textbf{O}\}\cap  \{O_m \neq O_*\}$, are treated as the reference objects, and only provide the cues of locations or relations to the target object.
	
	For the 3D modal input, each object is considered from the perspective of its 3D feature, semantic, size as well as relative position to the target object.
	Specifically, the object representation is computed as follows:
	\begin{align}
	F_{m}^{\text{3d}} &= \mathcal{T}_1([O_m^{f3d}; ~O_m^{cls}; ~W_1 O_m^{b3d}; ~W_2 O_m^{pe}]),
	\label{eq1}
	\end{align}
	where $[\cdot~;~\cdot]$ indicates the concatenation operation.
	$O_m^{f3d}$ is the output feature extracted by a 3D network, \eg, PointNet++~\cite{qi2017pointnet++}, and $O_m^{cls}$ is a one-hot vector for their predicted semantic class. 
	$O_m^{b3d}$ is the 3D bounding box of the object, consisting of the bounding box center $(x,y,z)$ and size $(w,h,l)$. 
	To a better object representation, we further design a positional encoding $O_m^{pe}$ for the $m$-th object as:
	
	\begin{equation}
	\begin{aligned}
	O^{pe}_m = [&O_m^x-O_*^x; O_m^y-O_*^y; O_m^z-O_*^z; \\
	&O^w_m/O^w_*; ~O^h_m/O^h_*; ~O^l_m/O^l_*].
	\label{eq1_1}
	\end{aligned}
	\end{equation}
	The first three elements in the positional encoding calculate the center offset between the target object and $m$-th object, and the others denote their relative size.
	Two learnable projection matrices $W_1$ and $W_2$ in Eqn.~\eqref{eq1} then transform the dimensions of $O_m^{b3d}$ and $O_m^{pe}$ to $d$.
	Finally, a transformation function $\mathcal{T}_1$ generates the final object feature $F_{m}^{3d}$ for the $m$-th object.
	
	Apart from 3D information in the multi-modal input, the corresponding 2D feature $O_m^{f2d}$ and 2D bounding box $O_m^{b2d}$ are introduced for the $m$-th object as follows:
	\begin{equation}
	\begin{aligned}
	F_{m}^{\text{multi}} = \mathcal{T}_2(&[O_m^{f3d}; ~O_m^{cls}; W_1 O_m^{b3d}; ~W_2 O_m^{pe}; \\&~O_m^{f2d}; ~W_3 O_m^{b2d}]).
	\label{eq2}
	\end{aligned}
	\end{equation}
	Concretely, for each object, its ground truth of the 3D bounding box is projected onto the original ScanNet videos~\cite{scannet} to obtain the corresponding 2D boxes.
	In each training step, an image is randomly selected from the video sequences to generate an extra input.
	Features in the 2D box area are extracted by the Faster-RCNN detector~\cite{ren2015faster} pre-trained on the Visual Genome~\cite{krishna2017visual} dataset, which are regarded as 2D features for the $m$-th instance, \ie, $O_m^{f2d}$.
	Finally, as shown in Eqn.~\eqref{eq2}, by applying linear and nonlinear transformations $W_3$ and $T_2$, a $d$-dimensional multi-modal representation $F_{m}^{\text{multi}}$ is generated for the $m$-th object.
	
	As shown in Figure~\ref{fig:fig2} (a), the multi-modal and 3D modal inputs have the same format, and each of them is a set of object features with the shape of $\mathbb{R}^{M\times d}$, in which the features of the target object and $M-1$ reference objects are entailed.
	For convenience, we denote the multi-modal input as $\textbf{F}^{\text{multi}} = \{F_{m}^{\text{multi}}\}_{m=1}^M$, and the 3D one as $\textbf{F}^{{\text{3d}}} = \{F_{m}^{\text{3d}}\}_{m=1}^M$.
	Then, these two inputs are fed to the teacher and student networks for the cross-modal knowledge transfer.

	\subsection{Baseline Model: TransCap}
	\label{sec:transformer}
	To apply the 3D and multi-modal object representation to our framework, we first introduce a baseline model, named TransCap, which employs Transformer~\cite{vaswani2017attention} structure to generate the descriptions of the target object.
	%
	%
	The student network in Figure~\ref{fig:fig2} (b) displays the architecture of TransCap. 
	It contains $L$ encoder layers and one decoder layer. 
	In each encoder layer, a self-attention mechanism is exploited to obtain a permutation invariant encoding for the input feature.
	Inspired by \cite{cornia2020meshed}, we design the self-attention operator $\mathrm{SA}(\cdot)$ as follows:
	\begin{equation}
	\begin{aligned}
	&\mathrm{SA}(\textbf{X}) = \mathrm{Attention}(\textbf{Q}, ~\textbf{K}, ~\textbf{V}),\\
	\textbf{Q} = W_q &\textbf{X}, ~\textbf{K} = [W_k \textbf{X}; ~M_k], ~\textbf{V} = [W_v \textbf{X}; ~M_v],
	\end{aligned}
	\end{equation}
	where $\textbf{X}\in \mathbb{R}^{M\times D}$ is a $D$-dimensional sequence, and $W_q$, $W_k$ and $W_v$ are matrices of learnable weights.
	Different from the traditional attention mechanism~\cite{vaswani2017attention}, the two persistent memory vectors $M_k$ and $M_v$ are appended to learn the prior knowledge.
	

	The captioning decoder is conditioned on previously generated words and features from the encoder layers to generate the next token.
	Specifically, it integrates the features from different encoder layers and performs the cross-attention on the generated tokens.
	\begin{equation}
	\mathrm{Decoder}(\hat{\textbf{X}}_l, \textbf{Y}) = \sum_{l=1}^{L}\alpha_l \odot \mathrm{CA}(\hat{\textbf{X}}_l, \textbf{Y}),
	\end{equation}
	where $\mathrm{CA}(\cdot, \cdot)$ stands for the encoder-decoder cross-attention~\cite{vaswani2017attention},
	computed using queries from the decoder output $\textbf{Y}$, and keys and values from the $l$-th layer encoder output $\hat{\textbf{X}}_l$.
	$\alpha$ are learnable weights having the same size as the
	cross-attention results.
	In this manner, TransCap takes a sequence of object features as the input, and generates the description for the target object.
	
	\begin{figure}[t]
		\begin{center}
			
			\includegraphics[width=0.95\linewidth]{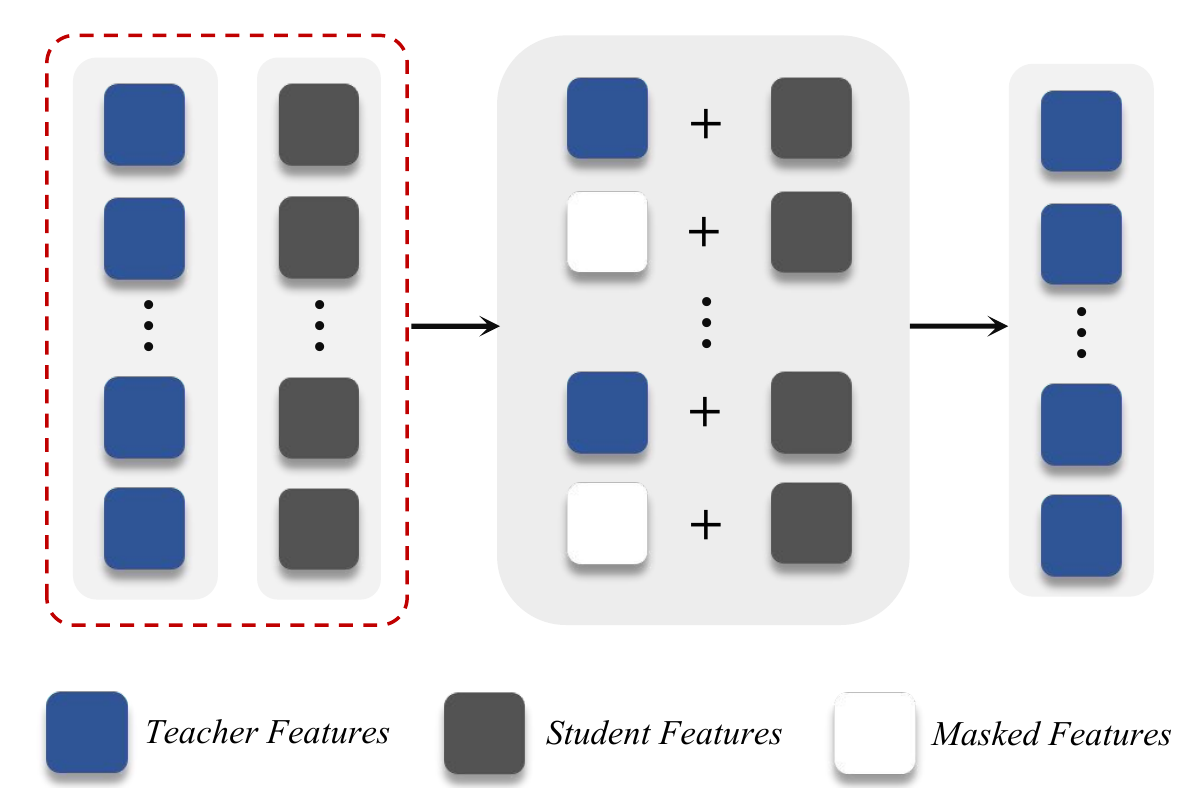}
			
			\caption{\textbf{Cross-Modal Fusion (CMF) Module.} 
				%
				%
				%
				It is designed to construct the interactionon from the student to the teacher network at the same encoder level, interacting the features of different modalities. During the training phrase, the features from the teacher network are randomly masked with a specific probability.
			}
			\label{fig:fig2_2}
		\end{center}	
	\end{figure}
	
	\subsection{Cross-Modal Fusion Module}
	\label{sec:interaction}
	The Cross-modal fusion (CMF) module enables cross-modal feature interaction between pure 3D and multi-modal feature representations.
	As shown in Figure~\ref{fig:fig2_2}, it is designed to construct interaction from the student network to the teacher at the same encoder level, thus building a bridge to fuse features between single and multiple modalities. 
	Moreover, to further enhance the ability of the student network to learn the multi-modal representation, we exploit a random mask on the features from the teacher network.
	Owing to this framework, the strengths of multi-modal representation can be assimilated to reinforce the student network via an end-to-end training protocol.
	Specifically, we element-wise add the student features with the masked teacher features.
	
	\begin{equation}
	\hat{\textbf{X}}_l^{*} = \hat{\textbf{X}}_l^{\text{student}} \oplus  ~\mathcal{I}(p)\hat{\textbf{X}}_l^{\text{teacher}},
	\label{eq3}
	\end{equation}
	where the $\hat{\textbf{X}}_l^{\text{student}}$ and $\hat{\textbf{X}}_l^{\text{teacher}}$ denote features from $l$-th encoder layer of student and teacher networks, respectively. The notation $\oplus$ means element-wise addition.
	The mask indicator $\mathcal{I}(p)$ is initialized with $1$ and has the probability of $p$ change to $0$.
	After that, we feed the fused features into the next encoder layer of the teacher network.
	It should be highlighted that, since our CMF module employs the single-directional connection from the student to the teacher, the teacher network can be discarded during inference, \ie, it introduces no extra computation for the student network.
	Moreover, various designs for the CMF module, including ablations, are shown in Section.~\ref{ablation}.
	
	\begin{table*}[t]
		\centering
		\small
		\caption{Comparison of 3D dense captioning results obtained by $\mathcal{X}$-Trans2Cap and previous methods using ground truth instances on ScanRefer and Nr3D datasets. We introduce the conventional captioning metrics, \ie, CIDEr~(C), BLEU-4~(B-4), METEOR~(M) and ROUGE~(R) for evaluation. The column `Extra 2D' means whether using extra 2D modality in the \textit{inference} phase. $\mathcal{X}$-Trans2Cap~(C) denotes the original model exploiting extra $\mathcal{L}_{\text{CIDEr}}$ loss in the final objective function.}
		\begin{tabular}{l|c|cccc|cccc}
			\toprule
			\multirow{2}[2]{*}{Method} & \multirow{2}[2]{*}{Extra 2D} & \multicolumn{4}{c|}{ScanRefer} & \multicolumn{4}{c}{Nr3D}  \\
			&       & C     & B-4   & M     & R     & C     & B-4   & M     & R \\\hline
			
			Scan2Cap~\cite{chen2021scan2cap} &   \xmark    & 65.79 & 38.54  & 28.81 & 61.93  & 63.36 & 32.07 & 28.92 & 64.56  \\
			Scan2Cap (Inst) &   \xmark    & 64.44 & 36.89  & 28.42 & 60.42  & 61.89 &  32.02 & 28.88 & 64.17 \\\hline
			TransCap &   \xmark    & 75.75 & 42.06 & 28.82 & 62.62 & 70.60 & 35.99 & 29.04 & 66.00  \\
			$\mathcal{X}$-Trans2Cap & \xmark & 87.09  & 44.12 & 30.67 & 64.37  & 80.02 & 37.90 & 30.48 & 67.64   \\
			$\mathcal{X}$-Trans2Cap (C) &   \xmark    & \textbf{97.17} & \textbf{45.70} & \textbf{31.23} & \textbf{64.23} & \textbf{81.44} & \textbf{39.08} & \textbf{30.79} & \textbf{68.15} \\\hline\hline
			Scan2Cap~\cite{chen2021scan2cap} &   \cmark & 67.95 & 41.49 & 29.23 & 63.66 & 64.13 & 32.98 & 29.75 & 65.24 \\
			Scan2Cap (Inst) &   \cmark  & 70.04 & 41.57 & 29.67 & 64.10 & 64.00 & 33.19 & 29.53 & 65.29 \\\hline
			TransCap &   \cmark    & 88.72 & 44.24 & 30.95 & 64.70 & 77.55 & 37.25 & 30.63 & 67.43   \\
			$\mathcal{X}$-Trans2Cap &   \cmark   & 89.73 & 44.25 & 31.00 & 64.50 & 85.38 & 39.52 & 31.23 & 68.18 \\
			$\mathcal{X}$-Trans2Cap (C) &   \cmark    & \textbf{106.11} & \textbf{49.07} & \textbf{32.25} &  \textbf{65.54} & \textbf{85.40} & \textbf{40.51} & \textbf{31.36} & \textbf{68.84} \\
			\bottomrule
		\end{tabular}%
		\label{tab:tab1}%
	\end{table*}%
	
	\subsection{Objective Function}
	\label{sec:loss}
	\noindent\textbf{Feature alignment loss.}
	Following a standard practice in knowledge transfer, we use Huber loss $\mathcal{L}_{\text{align}}$ (\ie, Smooth-L1 regression loss) to align decoder features between teacher and student networks.

	\noindent\textbf{Captioning loss.}
	As in the previous work~\cite{chen2021scan2cap}, we apply a conventional cross entropy loss function $\mathcal{L}_{\text{ce}}$ on the generated token probabilities in both teacher and student networks.
	Furthermore, to further boost the performance, we propose an enhanced version model $\mathcal{X}$-Trans2Cap~(C) by applying the CIDEr-D score~\cite{anderson2018bottom} $\mathcal{L}_{\text{CIDEr}}$ as reward.
	Following the previous work~\cite{cornia2020meshed}, we baseline the reward using the mean of the rewards rather than greedy decoding following previous methods~\cite{anderson2018bottom, rennie2017self}.

	\noindent\textbf{Total objective loss.}
	We combine all three loss terms linearly as our final objective loss function:
	\begin{equation}
	\mathcal{L} = \alpha \mathcal{L}_{\text{align}} + \beta \mathcal{L}_{\text{ce}} + \gamma \mathcal{L}_{\text{CIDEr}},
	\label{eq4}
	\end{equation}
	where $\alpha$, $\beta$ and $\gamma$ are the weights for each individual loss.
	To guarantee the loss terms are roughly of the same magnitude, we fine-tune weights on the validation split, and set those to $\alpha = 1$, $\beta = 1$, and $\gamma = 0.1$ empirically in the experiments.

	\subsection{Training and Inference Schemes}
	\label{sec:training}
	The black and red arrows in the Figure~\ref{fig:fig2} (b) illustrate the information flow of the $\mathcal{X}$-Trans2Cap for training and inference.
	It is noteworthy that teacher and student networks are trained from scratch.
	In the training phase, both networks are exploited (see the black and red arrows in Figure~\ref{fig:fig2} (b)), and CMF modules between corresponding encoder layers and feature alignment are conducted to enhance mutual representation.
	During the inference, if only the 3D modality exists, we only apply the student network (see the red arrow in Figure~\ref{fig:fig2} (b)).
	However, if the auxiliary 2D information is available as well, the stronger teacher framework will be exploited.
	In our experiment, we demonstrate that our architecture can both enhance the performance of teacher and student networks with and without additional modality.

	\section{Experiment}
	We compare our method with Scan2Cap and 2D baselines proposed in their paper. Extending from \cite{chen2021scan2cap}, we further compare all methods on Nr3D dataset~\cite{achlioptas2020referit3d}.
	More experiment results including subjective evaluation and ablations are in the supplementary material.
	
	\subsection{Datasets}
	\label{sec:data}
	\noindent\textbf{ScanRefer.}
	The ScanRefer dataset~\cite{chen2020scanrefer} annotates 800 3D indoor scenes in the ScanNet~\cite{scannet} dataset with 51,583 language queries. 
	It follows the official ScanNet splits and contains 36,665, 9,508, and 5,410 samples in train/val/test sets, respectively.
	Since the dataset is initially used in visual grounding and the labels of the test set are inaccessible, we follow the same setting as in \cite{chen2021scan2cap} to form the train and val sets for training and testing.

	\noindent\textbf{Nr3D.} The Natural Reference in 3D (Nr3D)~\cite{achlioptas2020referit3d} has the same train/val split as ScanRefer.
	It contains 41,503 queries annotated by Amazon Mechanical Turk (AMT) workers.
	Compared with ScanRefer dataset, Nr3D is more challenging since it does not contain the fixed or redundant sentence patterns, \ie, declarative sentences starting with ``\textit{this is}" or ``\textit{that is}".
	We do not compare our method on its counterpart, the Spatial Reference in 3D (Sr3D) dataset, since it is totally generated by the machine templates.
	
	\subsection{Tasks and Metrics}
	\noindent\textbf{Tasks.}
	In our experiment, we follow \cite{chen2021scan2cap} and design two protocols to evaluate the generated caption:
	\begin{itemize}
		\setlength{\itemsep}{0pt}
		\setlength{\parsep}{0pt}
		\setlength{\parskip}{0pt}
		\item \textit{Dense captioning with ground truth instances} (Oracle DC): In this setting, the point cloud of each instance is given. Then one needs to generate faithful captions based on their attribute information and spatial relationships.
		\item \textit{Dense captioning with 3D scans} (Scan DC): This setting is more challenging. One needs to detect objects from the 3D scans first and then generate captions for each object according to the detection results.
		
	\end{itemize}

	\noindent\textbf{Metrics.}
	In Oracle DC, we directly apply CIDEr~\cite{vedantam2015CIDEr}, BLEU-4~\cite{papineni2002bleu}, METEOR~\cite{banerjee2005meteor} and ROUGE~\cite{lin2004rouge} averagely on all instances as metrics. 
	For brevity, we simplify them as C, B-4, M and R, correspondingly.
	
	In Scan DC, to jointly measure the quality of generated captions and detected bounding boxes, we evaluate them by combining above metrics with Intersection-over-Union (IoU) scores between predicted bounding boxes and GT bounding boxes.
	Specifically, we follow ~\cite{chen2021scan2cap} and define the combined metrics as $m@k$IoU$=\frac{1}{N}\sum_{i=1}^N
	m_i u_i$, where $u_i \in \{0,1\}$ is set to 1 if the IoU score for the $i$-th box exceeds $k$, otherwise 0. We use $m$ to represent the above captioning metrics, \eg, CIDEr.
	$N$ is the number of detected object bounding boxes. 
	We also use mean average precision (mAP) thresholded by IoU as the object detection metric.
	
	%

	\begin{table*}[t]
		\centering
		\small
		\caption{Comparison of 3D dense captioning obtained by $\mathcal{X}$-Trans2Cap and previous methods, taking 3D Scans as the input on ScanRefer dataset. We average the scores of the above captioning metrics, which are with the IoU percentage between the predicted bounding box and the ground truth over 0.25 and 0.5, respectively. The `Extra 2D' means that whether using the extra 2D modality as above. {\textit{2D-3D Proj.}} and {\textit{3D-2D Proj.}} represent the method in~\cite{chen2021scan2cap}, \ie, 2D proposal mapped to 3D and 3D proposal projected to 2D images, respectively. {{`Proposals'}} shows the methods exploited to obtain 2D or 3D proposals.}
		\resizebox{\textwidth}{!}{
			\begin{tabular}{l|c|c|cccc|cccc|c}
				\toprule
				Method & \multicolumn{1}{l|}{Extra 2D} & Proposals & \multicolumn{1}{l}{{C@0.25}} & \multicolumn{1}{l}{{B-4@0.25}} & \multicolumn{1}{l}{{M@0.25}} & \multicolumn{1}{l}{{R@0.25}} & \multicolumn{1}{|l}{{C@0.5}} & \multicolumn{1}{l}{{B-4@0.5}} & \multicolumn{1}{l}{{M@0.5}} & \multicolumn{1}{l}{{R@0.5}} & \multicolumn{1}{|l}{{mAP@0.5}} \\\hline
				Scan2cap~\cite{chen2021scan2cap} &   \xmark    & VoteNet &  50.71  & 33.01  &   25.47    & 53.60 &  33.53     & 21.58  & 21.04 & 43.03 & 32.46 \\\hline
				TransCap &   \xmark    & VoteNet &  55.36 & 32.46 & 25.64 &    53.19   & 40.08 & 22.86 & 21.72 & 44.04 & 33.34  \\
				$\mathcal{X}$-Trans2Cap &   \xmark    & VoteNet & \textbf{58.81} & \textbf{34.17} & \textbf{25.81} & \textbf{54.10} & \textbf{41.52} & \textbf{23.83} & \textbf{21.90} & \textbf{44.97} & \textbf{34.68}  \\\hline\hline
				2D-3D Proj.~\cite{xu2015show} &   \cmark    & Mask R-CNN & 18.29 & 10.27 & 16.67 & 33.63 & 8.31  & 2.31  & 12.54 & 25.93 & 10.50 \\
				3D-2D Proj.~\cite{xu2015show} &    \cmark   & VoteNet & 19.73 & 17.86 & 19.83 & 40.68 & 11.47 & 8.56  & 15.73 & 31.65 & 31.83 \\
				Scan2cap~\cite{chen2021scan2cap} &   \cmark    & VoteNet & 56.82 & 34.18 & 26.29 & \textbf{55.27} & 39.08 & 23.32 & 21.97 & 44.78 & 32.21 \\\hline
				TransCap &   \cmark    & VoteNet & 60.04 & 35.04 & 26.27 & 54.46 & 43.12 & 24.25 & 22.15 & 44.72 & 34.34 \\
				$\mathcal{X}$-Trans2Cap & \cmark & VoteNet & \textbf{61.83} & \textbf{35.65} & \textbf{26.61} & {54.70} & \textbf{43.87} & \textbf{25.05} & \textbf{22.46} & \textbf{45.28} & \textbf{35.31} \\
				\bottomrule
		\end{tabular}}%
		\label{tab:tab2}%
	\end{table*}%
	\begin{figure*}[t]
		\centering\includegraphics[width=\linewidth]{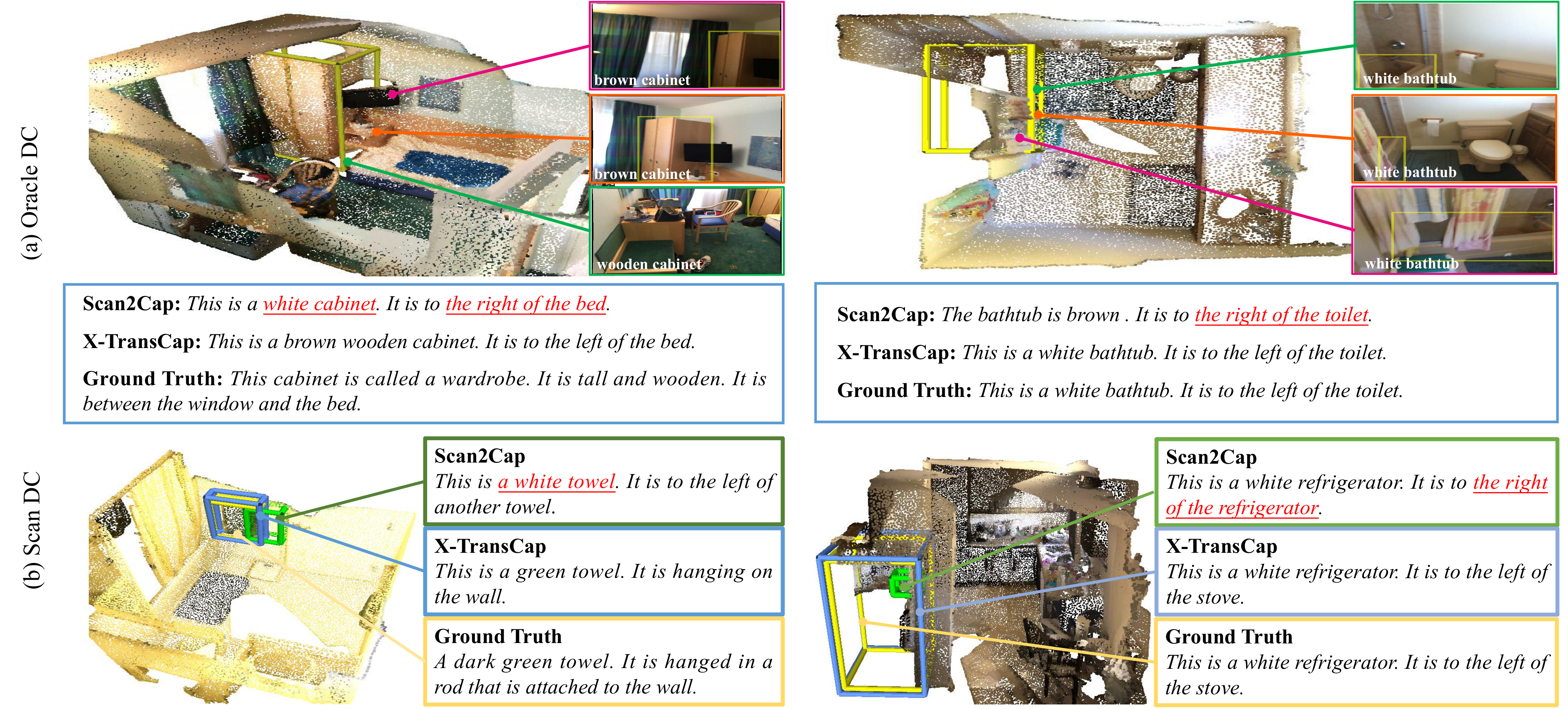}
		\vspace{.1cm}
		\caption{\textbf{Qualitative comparisons on ScanRefer dataset.} The upper illustrates the results applying ground truth instances (\ie, Oracle DC task), and the lower part presents the results using object detection (\ie, Scan DC task). Best viewed in color.}
		\label{fig:fig3}
		
	\end{figure*}
	
	\subsection{3D Dense Captioning Results}
	\noindent\textbf{Oracle dense captioning.}
	The results of Oracle DC task are displayed in Table~\ref{tab:tab1}.
	In the upper part, we compare results without the extra 2D input for inference.
	Scan2Cap and Scan2Cap (Inst) denote the methods that exploit ground-truth (GT) boxes and GT instances as input, respectively.
	Merely using the baseline model TransCap, we improve the captioning result by a large margin compared to Scan2Cap (+9.96 and +7.24 CIDEr points on ScanRefer and Nr3D). 
	Utilizing our cross-modal knowledge transfer training strategy further boost the performance on all the captioning metrics.
	Specifically, after using our teacher student framework, $\mathcal{X}$-Trans2Cap achieves 11.04 and 9.42 CIDEr improvement over TransCap on the ScanRefer and Nr3D datasets, where the performance on both datasets are about {\textbf{20}} CIDEr scores higher than those of Scan2Cap.
	The bottom part of Table~\ref{tab:tab1} illustrates the result using extra 2D input in both the training and inference phases.
	Though using the extra 2D input for inference, the performance of Scan2Cap is still inferior to that of our propsed $\mathcal{X}$-Trans2Cap only exploiting 3D modal input, let alone using multi-modal.
	Besides, $\mathcal{X}$-Trans2Cap is better than TransCap when both using the extra 2D input, especially on Nr3D (85.38 vs 77.55 CIDEr), which illustrates that training with student network can even improve the result of teacher network.
	Furthermore, with CIDEr-D score optimization, \ie, $\mathcal{X}$-Trans2Cap (C) model, the performance of captioning can be further improved.
	
	\noindent\textbf{Scan dense captioning.}
	In Table~\ref{tab:tab2}, we compare the result of Scan DC, which shows results without and with extra 2D input in the inference phase. The method for proposal generation is listed in the third column.
	Among these methods, {\textit{2D-3D Proj.}} and {\textit{3D-2D Proj.}} are two baseline methods proposed in \cite{chen2021scan2cap}.
	{\textit{2D-3D Proj.}} applies Mask R-CNN~\cite{he2017mask} to generate 2D proposals in images, where the corresponding 2D bounding boxes and features are fed into the description generation module~ \cite{chen2021scan2cap}.
	On the contrary, {\textit{3D-2D Proj.}} exploits VoteNet~\cite{qi2019deep} to extract 3D proposals, which are projected back to 2D images. Then the projected 2D proposals are finally adopted \cite{xu2015show} to generate captions.
	As shown in Table~\ref{tab:tab2}, 2D-based methods obtain lowest captioning scores, which reveals that they cannot directly handle the 3D dense captioning task.
	Though Scan2Cap achieves better results than these 2D based methods, it is much inferior to $\mathcal{X}$-Trans2Cap without the assistance of appealing 2D priors and dedicated network structure.
	Surprisingly, we observe that the detection performance of $\mathcal{X}$-Trans2Cap is improved as well, though there is no extra 2D input fed into the detector during training and testing.
	It confirms that our $\mathcal{X}$-Trans2Cap is not only capable of faithful caption generation, but also acquires the knowledge mining capacity within multi-modalities for more complex applications, \ie, digging out the information embedded into language description for 3D visual detection.
	The results of Scan DC on Nr3D dataset are illustrated in the supplementary.
	
	%
	
	\noindent\textbf{Visualization.}
	Figure~\ref{fig:fig3} presents the visualization results of $\mathcal{X}$-Trans2Cap, which demonstrates great improvements upon Scan2Cap for more faithful captions.
	Furthermore, we present the corresponding 2D counterparts within each 3D scene.
	Regarding 2D images, they can provide stronger texture and color information obviously when compared with sparse point cloud.
	
	\begin{table}[t]
		\centering
		\caption{{Comparison for knowledge transfer.} The results are obtained in Oracal DC on the ScanRefer dataset, where we compare both single-modal and cross-modal knowledge transfer methods.
		}
		\resizebox{\linewidth}{!}{
			\begin{tabular}{l|cccc}
				\toprule
				Method (Year) & \multicolumn{1}{c}{C} & \multicolumn{1}{c}{B-4} & \multicolumn{1}{c}{M} & \multicolumn{1}{c}{R} \\\hline
				Hinton \etal~\cite{hinton2015distilling} (2015) &   81.43    &    42.85   &   30.40    &   64.07 \\
				Huang \etal~\cite{huang2021revisiting} (2021) &   78.61    &    41.93   &   30.14    &   63.78 \\
				Pixel-to-point~\cite{liu2021learning}  (2021)&   77.82    &    41.98   &   29.42    &   62.79 \\
				2D SAT~\cite{yang2021sat}  (2021) & 80.13 & 41.13 & 30.00 & 63.16 \\\hline
				TransCap (baseline) &  75.75   & 42.06   &  28.82  & 62.62 \\
				$\mathcal{X}$-Trans2Cap (pre-trained) &    79.41   &   42.78    &   29.88    &  63.41 \\
				$\mathcal{X}$-Trans2Cap &  \textbf{87.09}   & \textbf{44.12}   &  \textbf{30.67}  & \textbf{64.37} \\
				\bottomrule
		\end{tabular}}
		\label{tab:kt}%
	\end{table}%

	\noindent\textbf{Comparison for knowledge transfer.}
	To further verify the effectiveness of our proposed method upon common teach-student architecture and other cross-modal manners, we compare $\mathcal{X}$-Trans2Cap with typical approaches of knowledge transfer in Table~\ref{tab:kt}.
	Among all the methods, Hinton \etal~\cite{hinton2015distilling} and Huang \etal~\cite{huang2021revisiting} are pure knowledge distillation designs, where the former is the pioneer for the research filed and the latter is newly proposed.
	As shown in the table, pure knowledge distillation manners cannot be directly adopted on the 3D DC scenario, and their improvement upon the baseline model is limited.
	Very recently, approaches~\cite{liu2021learning} and \cite{yang2021sat} adopt cross-modal knowledge transfer technique in the 3D tasks.
	The core idea of \cite{liu2021learning} is using extra 2D input to conduct 3D pre-training.
	We modify it by first training a TransCap with multi-modal input, then use its pre-trained parameters as initialization weights for pure 3D input training.
	For 2D semantic-assisted training (SAT)~\cite{yang2021sat}, it treats 2D features as additional tokens in (\ie, concatenated in sequence dimension) the same model, and then exploits an attention mask in Transformer layers. This mask only ignores the attention from 3D to the 2D.
	%
	However, both methods cannot boost the performance as no mutual enhancement is introduced.
	
	We also offer an offline distillation design, preparing a pre-trained teacher network before training the student network, called $\mathcal{X}$-Trans2Cap (pre-trained) in the bottom part of the Table~\ref{tab:kt}.
	It can be noticed that using pre-trained teacher network results in a performance drop of 7.38 CIDEr, which may result from the distribution gap between multi-modality data. 
	To the end, in the Table~\ref{tab:kt}, $\mathcal{X}$-Trans2Cap significantly performs better, which illustrates the effectiveness of the teacher-student framework and cross-modal fusion (CMF) module.

	\subsection{Analysis and Ablation Studies}
	\label{ablation}
	\noindent\textbf{Does knowledge transfer help?}
	As results shown in Table~\ref{tab:tab1} and \ref{tab:tab2},
	when we adopt 2D prior during the training phase ($\mathcal{X}$-Trans2Cap), it can greatly improve the performance upon the baseline model (TransCap).
	
	\noindent\textbf{Does our proposed components help?}
	To further verify the effectiveness of different components, we conduct the ablation studies in the Table~\ref{tab:tab4}.
	As shown in Table~\ref{tab:tab4}, model A is our baseline model (TransCap), and model B is our entire architecture of $\mathcal{X}$-Trans2Cap.
	The model C is the ablated architecture that discards the feature alignment loss $\mathcal{L}_{\text{align}}$.
	It can be found out that there is a great performance drop from 87.09 to 79.58 in terms of CIDEr metric. 
	Fortunately, due to the advantage of CMF module, it still has 3.83 improvement upon the baseline model.
	Similarly, the performance drop is appearing (-5.25 CIDEr) when removing the CMF module (model D).
	This result demonstrates that both framework architecture and CMF module play important roles in $\mathcal{X}$-Trans2Cap. 
	
	\noindent\textbf{How to design cross-modal fusion?}
	We illustrate the results from different designs of CMF module in the bottom part of the Table~\ref{tab:tab4}.
	On the one hand, discarding the random mask hampers the caption results, as shown in model~F.
	On the other hand, exploiting more complicated operations such as concatenation and attention mechanism cannot effectively improve the performance.
	There is only a slight improvement on the metrics of BLEU-4 and Rough for adopting attention.
	However, it will greatly increase the model complexity while make the CIDEr decline.
	

	\begin{table}[t]
		\centering
		\caption{{Ablation study for applying different knowledge transfer designs.} The results are obtained in Oracal DC on the ScanRefer dataset. The upper part shows ablated results without specific components, and the lower illustrates results of CMF designs.
		}
		\resizebox{\linewidth}{!}{
			\begin{tabular}{l|l|cccc}
				\toprule
				& Design & \multicolumn{1}{c}{C} & \multicolumn{1}{c}{B-4} & \multicolumn{1}{c}{M} & \multicolumn{1}{c}{R} \\\hline
				A& TransCap (baseline) &  75.75   & 42.06   &  28.82  & 62.62 \\ 
				B&$\mathcal{X}$-Trans2Cap &  \textbf{87.09}   & {44.12}   &  \textbf{30.67}  & {64.37} \\\hline
				C&w/o $\mathcal{L}_{align}$ & 79.58 & 41.47 & 30.07  & 63.59  \\
				D&w/o CMF & 80.54 & 43.15 & 30.19 & 63.59 \\\hline
				E& Concatenation &  79.87 &  43.28 & 30.22& 64.88 \\
				F& w/o Randon Mask & {85.36}   & {42.57}   &  {30.52}  & {64.18}   \\
				G&  Attention~\cite{vaswani2017attention}  & 80.85 & \textbf{44.75} & 30.45 & \textbf{64.91} \\
				\bottomrule
		\end{tabular}}
		\label{tab:tab4}%
	\end{table}%

	\section{Conclusion}
	
	In this work, we propose an enhanced 3D dense captioning method via cross-modal knowledge transfer, named \textbf{$\mathcal{X}$-Trans2Cap.}
	By designing the network architecture and knowledge distillation method carefully, our $\mathcal{X}$-Trans2Cap outperforms previous methods by a large margin on multiple datasets with more faithful captions.
	We believe that our work can be applied to a wider range of  3D vision and language scenarios, and provide a solution to the comprehension of 3D scenes with severe texture details missing, \ie, leveraging the 2D priors and the cross-modal knowledge transfer to improve the performance.
	
	\section*{Acknowledgment}
	
	{\noindent This work was supported in part by NSFC-Youth  61902335,  by Key Area R\&D Program of Guangdong Province with grant No.2018B030338001, by the National Key R\&D Program of China with grant No.2018YFB1800800,  by Shenzhen Outstanding Talents Training Fund, by Guangdong Research  Project No.2017ZT07X152, by Guangdong Regional Joint Fund-Key Projects 2019B1515120039,  by the  NSFC 61931024\&81922046, by helixon biotechnology company Fund and CCF-Tencent Open Fund.}
	\newpage
	{
		\bibliographystyle{ieee_fullname}
		\bibliography{camera_ready.bbl}
	}
	\newpage
	\noindent\title{\large\textbf{{Supplementary Material}}}S

	\setcounter{section}{0}
	\setcounter{figure}{0}
	\setcounter{table}{0}
	\renewcommand\thesection{\Alph{section}}
	
	
	\section{Overview}
	In this supplementary material, we illustrate the implementation details, the efficiency of the model and the results of subjective evaluation in Section~\ref{impl}, Section~\ref{speed} and Section~\ref{subjective}, respectively.
	After that, we provide more experiments of Scan Dense Captioning (DC) on Nr3D dataset in Section~\ref{sec:exp}. 
	Then we discuss the effectiveness of each attribute in instance representation in Section~\ref{sec:rep}.
	
	\section{Implementation Details}
	\label{impl}
	In our experiment, we adopt the PointNet++ to generate 3D object features ($O_m^{f_{\text{3d}}}$) in Oracle DC, and applies proposals' features from VoteNet in the Scan DC.
	Furthermore, in the test with Oracle DC, we use ground truth category as $O_m^{cls}$ while adopting the predicted results from detector in Scan DC task.
	We train the network for 30 epochs by using Adam optimizer with a batch size of 32. 
	The probability of random mask in CMF module is set as 0.2 when achieving the best, and it does not greatly change the result.
	It should be noted that both teacher and student networks are trained from scratch.
	The learning rate is initialized as 0.0005 with the decay as 0.1 for every 10 epochs.
	Experiments are conducted on RTX2080Ti GPUs. 
	
	\section{Running Time Evaluation}
	\label{speed}
	We investigate the running time of our model in this section.
	Table~\ref{tabs:tab1} shows the number of parameters and inference time of per scan in Oracle DC setting.
	\X (3D) can speed up more than \textbf{20$\times$} compared with its baseline and \X using extra 2D modality.
	%
	%
	%
	%


	\section{Subjective Evaluation}
	\label{subjective}
	We conduct a subjective evaluation with three volunteers on randomly selected 100 descriptions generated by Scan2Cap and $\mathcal{X}$-Trans2Cap with \textit{Oracle
		DC} setting on ScanRefer datasets. The subjective evaluation results are shown in Table~\ref{tab:tab3}. 
	In practice, each volunteer is asked to manually check whether the descriptions correctly reflect two aspects of the object: object color attributes and spatial relations in local environment. 
	As observed from Table~\ref{tab:tab3}, $\mathcal{X}$-Trans2Cap can generate more faithful captions regarding the attributes and spatial relationships. 
	
	\section{Scan Dense Captioning on Nr3D}
	\label{sec:exp}
	In Table~\ref{tab:sup_tab1}, we compare the results of Scan DC on Nr3D, including the results without and with extra 2D input in the inference phase. All methods exploit the same network, \ie, VoteNet, to generate proposals.
	{\textit{3D-2D Proj.}} projects proposals back to 2D images and captions in a 2D manner.
	However, it achieves the lowest captioning scores, which reveals that it cannot directly handle the 3D dense captioning task.
	Though Scan2Cap achieves better results than \textit{3D-2D Proj.}, it also cannot generate faithful captioning results.
	Not surprisingly, $\mathcal{X}$-Trans2Cap obtains the highest score in all metrics.
	Specifically, it not only gains a +2.9 improvement in CIDEr@0.25 score upon baseline TransCap, but also achieves +5.5 boost over Scan2Cap. 
	Finally, the experiment also confirms that our $\mathcal{X}$-Trans2Cap can improve 3D visual detection as well.
	%

	\begin{table}[t]
		\small
		\caption{{The complexity analysis between \X using both 3D and 2D inputs and only 2D input.} Here \underline{underline} correspond to the time and parameters for the 2D feature extractor.\vspace{-.6cm}}
		\begin{center}
			
			\begin{tabular}{l|c|cc}
				\hline
				Method & 2D & \#Param (M) &  Inference (s) \\\hline
				TransCap & \xmark & 19.9 & 0.4  \\
				TransCap & \cmark &  \underline{60.0}+19.9 & \underline{8.1}+0.4   \\
				\X & \xmark & \textbf{19.9} & \textbf{0.4}   \\
				\X & \cmark & \underline{60.0}+38.8 & \underline{8.1}+0.9 \\
				\hline
				
			\end{tabular}
		\end{center}
		
		\label{tabs:tab1}
	\end{table}
	
	\begin{table}[t]
		\centering
		\footnotesize
		\caption{
			{Subjective evaluation in Oracle DC setting.}  We measure the accuracy of two aspects (object colors and spatial relations) in the generated captions. 
		}
		\resizebox{\linewidth}{!}{
			\begin{tabular}{l|c|cc}
				\toprule
				Design & Extra 2D &\multicolumn{1}{c}{Attribute} & \multicolumn{1}{c}{Relation} \\\hline
				Scan2Cap & \xmark  & 61.82   &   66.86    \\
				$\mathcal{X}$-Trans2Cap & \xmark &  \textbf{68.73} (+6.91)   &   \textbf{75.54} (+8.68)   \\\hline
				Scan2Cap & \cmark  & 64.21 &   69.00   \\
				$\mathcal{X}$-Trans2Cap & \cmark &  \textbf{70.12} (+5.91)   &   \textbf{78.97} (+9.97)    \\
				
				\bottomrule
			\end{tabular}
		}
		\label{tab:tab3}%
	\end{table}%
	

	\begin{table*}[t]
		\centering
		\footnotesize
		\caption{Comparison of 3D dense captioning obtained by $\mathcal{X}$-Trans2Cap and previous methods, taking 3D Scans as the input on Nr3D dataset. We average the scores of the above captioning metrics, which are with the IoU percentage between the predicted bounding box and the ground truth over 0.25 and 0.5, respectively. The `Extra 2D' means that whether using the extra 2D modality as above. {\textit{3D-2D Proj.}} represents the method in Scan2Cap, \ie, 3D proposal projected to 2D images. {{`Proposals'}} shows the methods exploited to obtain 2D or 3D proposals.}
		\resizebox{\textwidth}{!}{
			\begin{tabular}{l|c|c|cccc|cccc|c}
				\toprule
				Method & \multicolumn{1}{l|}{Extra 2D} & Proposals & \multicolumn{1}{l}{{C@0.25}} & \multicolumn{1}{l}{{B-4@0.25}} & \multicolumn{1}{l}{{M@0.25}} & \multicolumn{1}{l}{{R@0.25}} & \multicolumn{1}{|l}{{C@0.5}} & \multicolumn{1}{l}{{B-4@0.5}} & \multicolumn{1}{l}{{M@0.5}} & \multicolumn{1}{l}{{R@0.5}} & \multicolumn{1}{|l}{{mAP@0.5}} \\\hline
				Scan2cap  & \xmark    & VoteNet & 41.76 & 24.12 & 24.98 & 55.79 & 23.70 & 14.88 & 20.95 & 47.50& 32.17 \\
				TransCap &   \xmark    & VoteNet &   44.32 & 25.63 & 25.25 & 55.69 & 27.24 & 17.76 & 21.60 & 49.16 & 34.09 \\
				$\mathcal{X}$-Trans2Cap &   \xmark    & VoteNet & \textbf{47.26} & \textbf{27.38} & \textbf{25.45} & \textbf{56.28}  & \textbf{30.96} & \textbf{18.70} & \textbf{22.15} & \textbf{49.92} & \textbf{34.13} \\\hline
				3D-2D Proj. &    \cmark   & VoteNet & 8.57 & 8.49 & 18.83 & 44.95 &3.93 &4.21 &16.68 & 41.24 & 31.83 \\
				Scan2cap &   \cmark    & VoteNet & 42.24 & 24.43&25.07&55.88&24.10&15.01&21.01&47.95&32.21\\
				TransCap &   \cmark    & VoteNet & 45.06 & 25.79 & 25.22 & 55.55 & 33.45 & 19.09 & 22.24 & {50.00} &  33.71 \\
				$\mathcal{X}$-Trans2Cap & \cmark & VoteNet & \textbf{51.43} & \textbf{27.62} & \textbf{25.75} & \textbf{56.46} & \textbf{33.62} & \textbf{19.29} & \textbf{22.27} & \textbf{50.00} & \textbf{34.38}  \\
				\bottomrule
		\end{tabular}}%
		\label{tab:sup_tab1}%
	\end{table*}%

	
	
	

	\begin{table*}[t]
		\centering
		\footnotesize
		\caption{{Ablation study for applying different instance representation designs.} The results are obtained in Oracal DC on the ScanRefer dataset. The upper part shows ablated results for different student input design, and the lower illustrates results using specific input for teacher network.
		}
		\resizebox{\linewidth}{!}{
			\begin{tabular}{c|cccccc|cccc|cccc}
				\toprule
				&  \multicolumn{6}{c|}{Teacher Network} & \multicolumn{4}{c|}{Student Network} & \multicolumn{4}{c}{Metrics} \\
				Model & $O^{f3d}$ & $O^{cls}$ & $O^{b3d}$ & $O^{pe}$ & $O^{f2d}$ & $O^{b2d}$ & $O^{f3d}$ & $O^{cls}$ & $O^{b3d}$ & $O^{pe}$  & \multicolumn{1}{c}{C} & \multicolumn{1}{c}{B-4} & \multicolumn{1}{c}{M} & \multicolumn{1}{c}{R} \\\hline
				A& \cmark & \cmark & \cmark &\cmark &\cmark &\cmark &\cmark &\cmark &\cmark &\cmark & \textbf{87.09}   & \textbf{44.12}   &  \textbf{30.67}  & {64.37}\\
				B& \cmark & \cmark & \cmark &\cmark &\cmark &\cmark &\cmark &\color{red}{\xmark} &\cmark &\cmark & 70.41 & 39.98 & 28.70 & 62.09\\
				C& \cmark & \cmark & \cmark &\cmark &\cmark &\cmark &\cmark &\cmark &\color{red}{\xmark} &\cmark &82.99  & 43.39 & 30.22 & \textbf{64.57}\\
				D& \cmark & \cmark &\cmark  & \cmark &\cmark &\cmark &\cmark &\cmark &\cmark &\color{red}{\xmark} & 33.52  & 35.67 & 26.33 & 61.78\\ \hline
				E& \color{red}{\xmark} & \cmark & \cmark &\cmark &\cmark &\cmark &\cmark &\cmark &\cmark &\cmark & 86.71 & 43.92 & {30.54} & 64.32\\
				F&  \cmark & \cmark & \cmark &\cmark &\color{red}{\xmark} &\cmark &\cmark &\cmark &\cmark &\cmark & 84.23 & 43.43 & 30.24 & 64.33\\
				G& \cmark & \cmark & \cmark &\cmark &\cmark &\color{red}{\xmark} &\cmark &\cmark &\cmark &\cmark & 83.85 & 43.12 & 30.24 & 64.52\\
				\bottomrule
		\end{tabular}}
		\label{tab:sup_tab2}%
	\end{table*}%
	
	\section{Analysis and Ablation Studies}
	\label{sec:rep}
	We further conduct an ablation study on different instance representation designs as shown in Table~\ref{tab:sup_tab2}, where the upper part and lower part show the specific designs in teacher and student network, respectively.
	
	\noindent\textbf{Does object class help?}
	From the results of model B in Table~\ref{tab:sup_tab2}, it can be found out that there is a dramatic drop in metric of CIDEr, from 87.09 to 70.41, when we discard the object class $O^{cls}$. 
	Thus, it shows that $O^{cls}$ is an important attribute for instance representation.
	Note that the ablated model B is still +6 CIDEr higher than that of Scan2Cap.
	
	\noindent\textbf{Does 3D bounding box help?}
	As shown in results of model C in Table~\ref{tab:sup_tab2}, removing the 3D bounding box $O^{b3d}$ will not cause a large performance drop, only -4.1 CIDEr from 87.09 to 82.99.
	This result reflects that $\mathcal{X}$-Trans2Cap utilizes 3D object spatial coordinates to generate captions.
	
	\noindent\textbf{Does positional encoding help?}
	The result of model D demonstrates a tremendous performance decrease in metric of CIDEr when positional encoding $O^{pe}$ is not exploited, where the model can only obtain 33.52 in metric of CIDEr.
	Since our model only chooses one object as the target object and the remaining ones will be regarded as reference objects, positional encoding helps the model to identify the target one.
	Without its help, the network can hardly work.

	\noindent\textbf{Does 2D input help?}
	The lower part of Table~\ref{tab:sup_tab2} describes the effectiveness of different attributes in the teacher network.
	There are three conclusions can be obtained:
	\textbf{1)} Discarding 3D features $O^{f3d}$ in teacher network barely hampers the performance (model E).
	This is because the 3D features also exist in the input of the student network.
	\textbf{2)} Utilizing the pre-trained network to extract 2D features is not necessary (model F).
	The result of model F shows that even if we only exploit the information of 2D bounding box, there is only an about -2 CIDEr drop for the caption results.
	\textbf{3)} The 2D bounding box information $O^{b2d}$ seems to play a more important role compared with 2D features $O^{f2d}$ (see the model G).
	Without using $O^{b2d}$, the model only obtains 83.85 CIDEr, and this result is even 0.4 lower than that of model F (without using $O^{f2d}$).
	Such results also emphasize the capability of  $\mathcal{X}$-Trans2Cap in real-world applications, \ie, without pre-trained 2D network, only utilizing the 2D bounding box information can still greatly boost the captioning performance.

\end{document}